\documentclass[10pt,twocolumn,letterpaper]{article}

\usepackage{iccv}
\usepackage{times}
\usepackage{epsfig}
\usepackage{graphicx}
\usepackage{amsmath}
\usepackage{amssymb}
\usepackage{booktabs}
\usepackage[T1]{fontenc}
\usepackage{dutchcal}
\usepackage{multirow}
\usepackage{subcaption}
\usepackage[pagebackref=true,breaklinks=true,letterpaper=true,colorlinks,bookmarks=false]{hyperref}
\usepackage[misc]{ifsym}

\usepackage[capitalize]{cleveref}
\iccvfinalcopy % *** Uncomment this line for the final submission

\def\iccvPaperID{9760} % *** Enter the ICCV Paper ID here

\ificcvfinal\fi

\begin{document}

\title{Multi-Graph Convolution Network for Pose Forecasting}

\author{Hongwei Ren\\
Polytechnic Institute\\
Zhejiang University\\
{\tt\small rhw@zju.edu.cn}
\and
Yuhong Shi\\
Polytechnic Institute\\
Zhejiang University\\
{\tt\small shi.yh@zju.edu.cn}
% Additional authors and addresses can \microsoft\meshgraphormerbe added with ``\and'',
\and
Kewei Liang\textsuperscript{(\Letter)}\\
School of Mathematical Sciences\\
Zhejiang University\\
{\tt\small matlkw@zju.edu.cn}
}
\maketitle

%%%%%%%%% ABSTRACT
\begin{abstract}
  Recently, there has been a growing interest in predicting human motion,
  which involves forecasting future body poses based on observed pose sequences. 
  This task is complex due to modeling spatial and temporal relationships. 
  The most commonly used models for this task are autoregressive models, 
  such as recurrent neural networks (RNNs) or variants, and Transformer Networks. 
  However, RNNs have several drawbacks, such as vanishing or exploding gradients.
  Other researchers have attempted to solve the communication problem in the spatial dimension by 
  integrating Graph Convolutional Networks (GCN) and Long Short-Term Memory (LSTM) models. 
  These works deal with temporal and spatial information separately,
  which limits the effectiveness.
  To fix this problem, we propose a novel approach called the multi-graph convolution network (MGCN) 
  for 3D human pose forecasting. 
  This model simultaneously captures spatial and temporal information 
  by introducing an augmented graph for pose sequences.
  Multiple frames give multiple parts, joined
  together in a single graph instance.
  Furthermore,
  we also explore the influence 
  of natural structure and sequence-aware attention to our model.
  In our experimental evaluation of the large-scale benchmark datasets, 
  Human3.6M\cite{Ionescu2014Human36MLS}, AMSS\cite{Mahmood2019AMASSAO} and 3DPW\cite{Marcard2018RecoveringA3},
  MGCN outperforms 
  the state-of-the-art in pose prediction (See \Cref{tab:experiment} and \Cref{fig:line_fig1}). 
\end{abstract}

\begin{figure}
  \begin{subfigure}{1\linewidth}
    % \fbox{\rule{0pt}{2in} \rule{.9\linewidth}{0pt}}
    \includegraphics[width=1\linewidth]{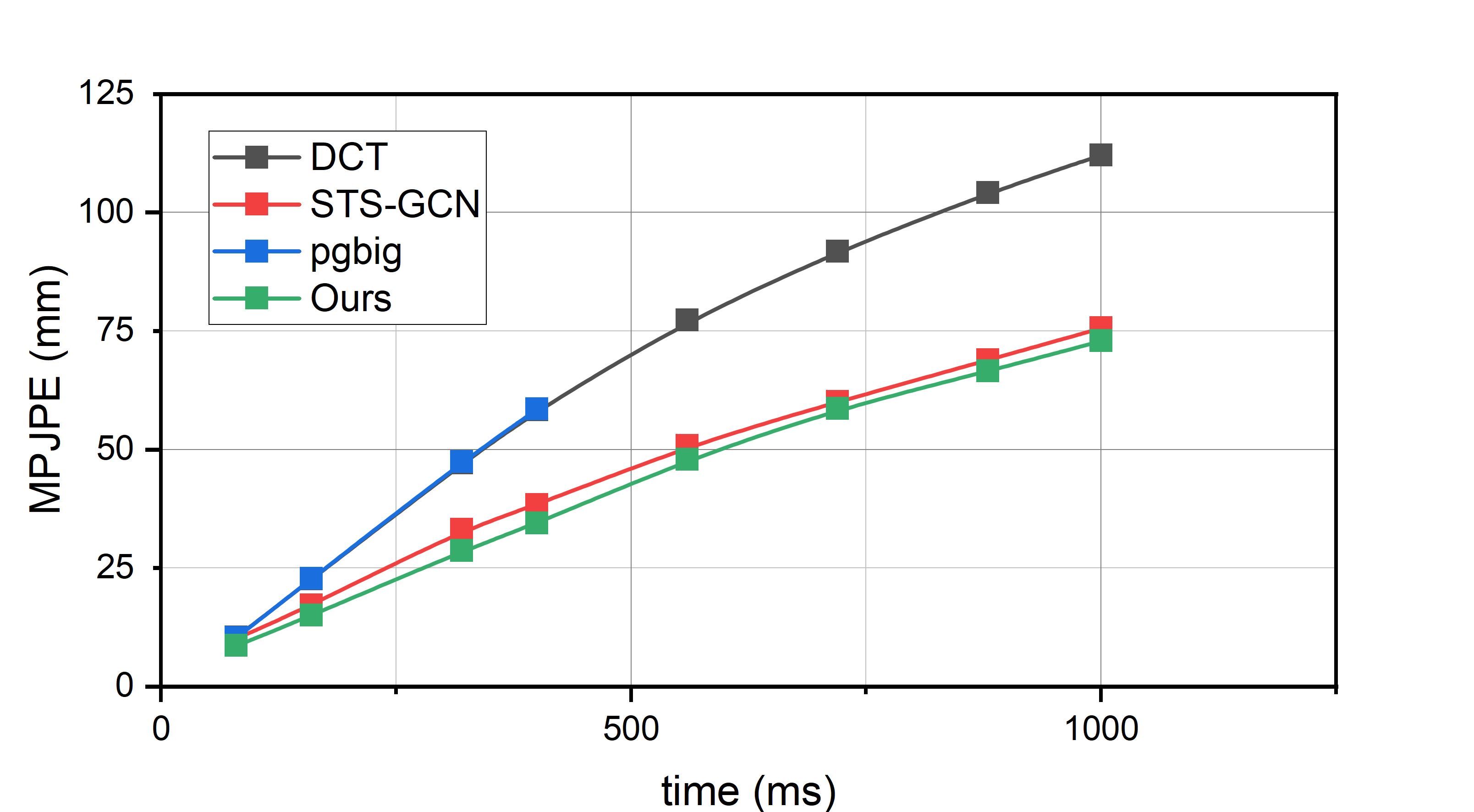}
    \caption{}
  \end{subfigure}
  \begin{subfigure}{1\linewidth}
    % \fbox{\rule{0pt}{2in} \rule{.9\linewidth}{0pt}}
    \includegraphics[width=1\linewidth]{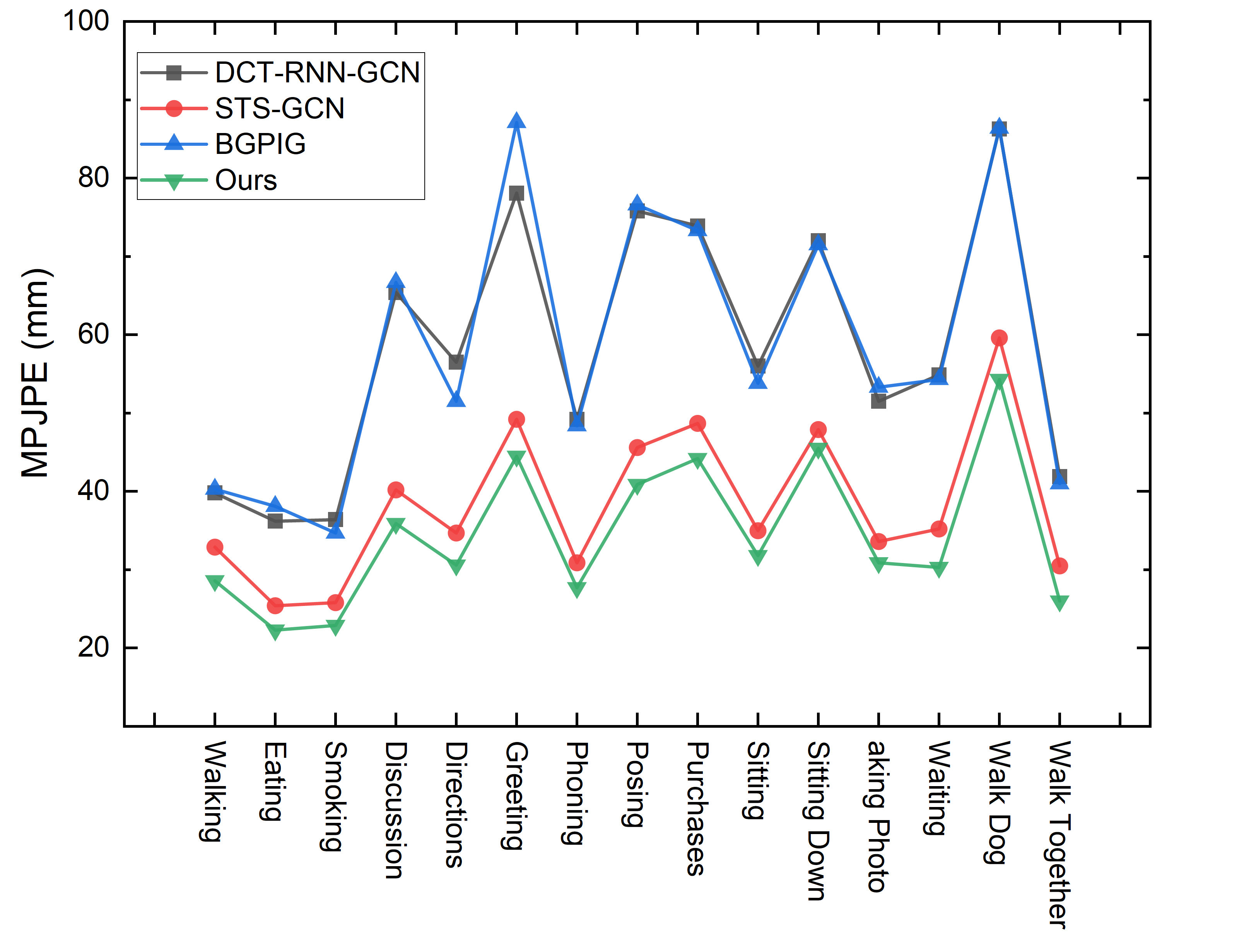}
    \caption{}
  \end{subfigure}
  \caption{(a) Performance of various methods on Human3.6M. The figure shows that our method outperforms others and achieves 
  state-of-the-art. (b) Performance of different actions on a 400 ms prediction task.}
\label{fig:line_fig1}
\end{figure}
%%%%%%%%% BODY TEXT
\section{Introduction}
\label{sec:intro} 
Human pose forecasting is a critical task that involves modeling the complete structured sequence of joints. 
It has various applications in human-robot interaction systems \cite{Koppula2013AnticipatingHA}, autonomous 
driving \cite{Paden2016ASO}, and healthcare \cite{Troje2002DecomposingBM}.
Many researchers \cite{chiu2019action,mao2020history,mao2019learning} use recurrent neural networks (RNNs) 
or variants to tackle this 
task due to the sequential nature of the data. However, RNNs have several drawbacks. Firstly, they are 
prone to vanishing or exploding gradients \cite{Pascanu2013OnTD}. 
Secondly, RNN-based methods often suffer from severe discontinuities between adjacent output frames, 
leading to poor global smoothing of the output sequence \cite{Li2018ConvolutionalST, Martinez2017OnHM}. 
To tackle these challenges,
Graph Convolutional Networks (GCN) have been introduced.
GCNs extract features from key point graphs 
and have achieved state-of-the-art performance on several benchmark datasets, as reported in 
\cite{sofianos2021space}.

% This paper proposes an approach to pose forecasting 
% which simultaneously models spatial and temporal body
% joints. Specifically, Graph Convolutional networks are 
% employed: a given time instant gives a pose, which is part of
% the graph. Multiple frames give multiple parts, joined 
% together in a single graph instance. 
% We also explore the influence of structure of human to GCN and
% further demonstrate that natural 
% human skeletal connections play an 
% important role in the recognition of 
% human movements\cite{Yan2018SpatialTG,sofianos2021space,Jain2016StructuralRNNDL,Btepage2017DeepRL}. 

% To the best of my knowledge,
Song \etal \cite{Song2020SpatialTemporalSG} utilize the connection between 
different frames for traffic management.
Inspired by this,
we propose an approach to pose forecasting 
which simultaneously models spatial and temporal body
joints. Specifically, Graph Convolutional networks are 
employed: a given time instant gives a pose, which is part of
the graph. Multiple frames give multiple parts, joined 
together in a single graph instance. 
% Additionally,
% Inspired by the autoregressive 
% mechanism and the decoder of Transformer \cite{Vaswani2017AttentionIA},
% we develop a masked attention mechanism to enable the model to 
% express the human pose in frame $t$ solely using the feature prior to $t$.
In addition, we have taken inspiration from the autoregressive mechanism and decoder of the 
Transformer \cite{Vaswani2017AttentionIA} to create a masked attention mechanism. 
This mechanism allows our model to accurately represent the human pose in frame $t$ 
using only the feature information prior to frame $t$.

\begin{figure*}
  \centering
     \includegraphics[width=0.8\linewidth]{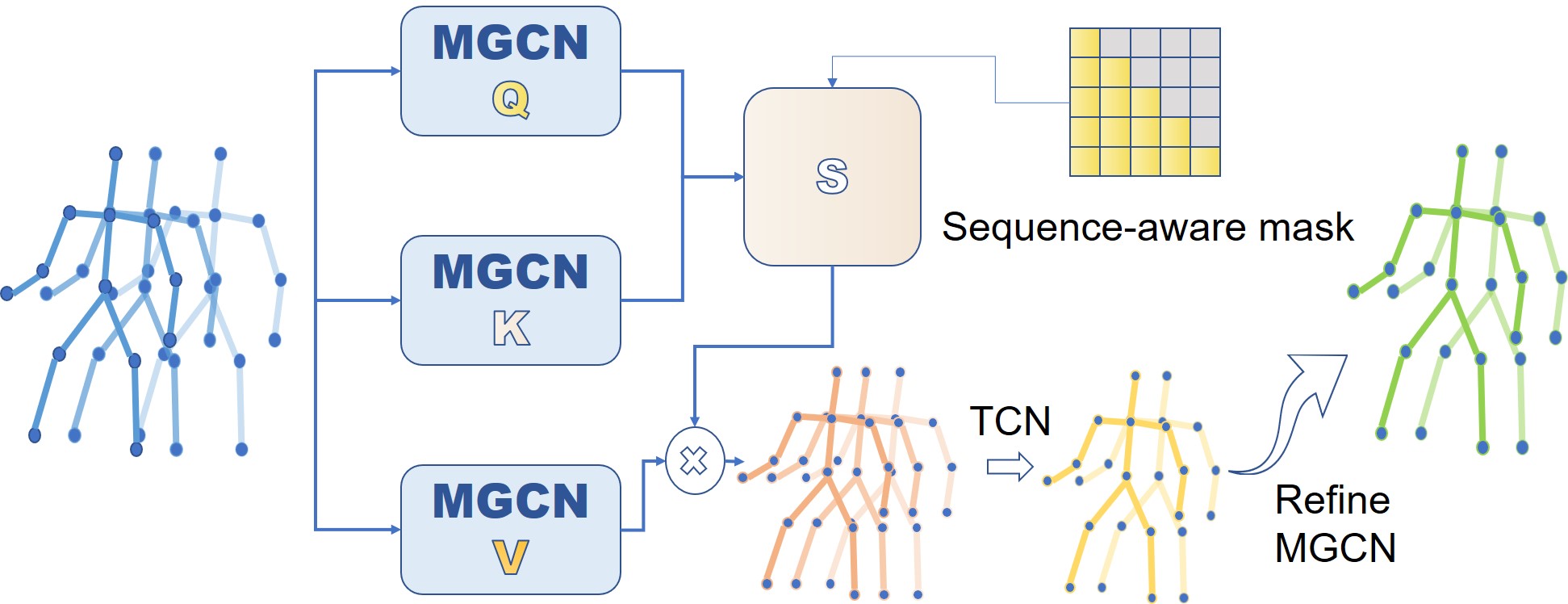}
  
     \caption{The complete pipelines of the proposed model. 
     The vectors Q, K, and V are all produced by the MGCN. %Network. 
     After the masked attention, 
     TCN is applied as a time sequence generator, 
     aligning the number of frames the model will predict.
     Finally, we use another MGCN module to refine the results.
     }
  \label{fig:label1}
  \end{figure*}

\Cref{fig:label1} and \Cref{fig:label2} illustrate the pipeline of our proposed model and 
the network architecture of MGCN respectively.
Firstly, we generate vectors Q, K, and V using MGCN. We
adopt a modified attention mechanism and add a mask to the S to perceive the sequences, 
which we refer to as "sequence-aware attention".
Next, we use temporal convolutional networks (TCNs \cite{Gehring2017ConvolutionalST,Bai2018AnEE}) 
as a time sequence generator to align the number of frames to predict.
Finally, we add a refinement process after the generator to achieve 
a smoother representation of the output frames.

The main contributions of this paper are summarized as 
follows:

$\bullet$ A novel multi-graph convolutional network with 
spatial-temporal reception is proposed.

$\bullet$ We propose two strategies on attention to make the 
model aware of the sequence, which achieve state-of-the-art at different tasks. 

$\bullet$ We experimentally prove the importance of natural links for GCN.

$\bullet$ The proposed model of MGCN significantly exceeds 
the state-of-the-art in Human3.6M, 3DPW and AMASS benchmarks.

% Our model largely outperforms the state-of-the-art 
% in a complex, recent, and large-scale benchmark Human3.6M.
%-------------------------------------------------------------------------

%-------------------------------------------------------------------------
\section{Related Work}

Considering whether graph convolution networks are used or not, 
we divide the recent works related to human pose forecasting into two parts.

\textbf{No-GCN methods} 
RNN \cite{chiu2019action,mao2020history,mao2019learning} or variants such as 
gated recurrent units (GRU) \cite{Yuan2020DLowDL} 
or long short-term memory networks (LSTM)
\cite{Yuan2019EgoPoseEA} have been widely adopted in human pose forecasting. 
% These techniques are flexible 
% but insufficient for long-term prediction and have main drawbacks, 
% such as training difficulties \cite{Pascanu2013OnTD}  
% and discontinuities between adjacent output frames 
% \cite{Li2018ConvolutionalST, Martinez2017OnHM}.
These flexible techniques have limitations for long-term 
prediction and can be challenging to train \cite{Pascanu2013OnTD}. 
Discontinuities between adjacent output frames may also arise \cite{Li2018ConvolutionalST, Martinez2017OnHM}.
Feed-forward approaches have been studied as an 
alternative way to solve the discontinuity problem.
Btepage \etal \cite{Btepage2017DeepRL} propose to treat a recent
pose history as input to a fully-connected network and introduce 
different strategies to encode additional temporal
information via convolutions and spatial structure by exploiting the kinematic tree. 
Similarly Diller \etal \cite{Diller2020ForecastingC3} 
also use previous joint prediction as input to automatic regress. 
Many excellent performances are attained with convolutional layers 
\cite{Li2018ConvolutionalST,mao2019learning} on the temporal 
dimension, which are known as temporal convolutional networks
(TCNs) \cite{Bai2018AnEE,Luo2018FastAF}. 
Here, we adopt TCN only for aligning the time dimension due to 
its performance. 
Besides, attention technique are also used to get a better understanding of temporal or spatial 
relations\cite{mao2019learning, Cai2020LearningPJ, Tang2018LongTermHM}. 
% Btepage et al. \cite{Btepage2017DeepRL} proposed a method that uses a recent history of poses as input 
% to a fully-connected network and introduces different strategies for encoding additional temporal 
% information through convolutions and exploiting the kinematic tree's spatial structure. 
% Diller et al. \cite{Diller2020ForecastingC3} also use previous joint predictions as input to an automatic 
% regressor. Many researchers have achieved excellent performance by using temporal convolutional 
% networks (TCNs) \cite{Bai2018AnEE,Luo2018FastAF}, which use convolutional layers on the temporal dimension. 
% In this work, we use TCNs for aligning the time dimension due to their superior performance. 
% Additionally, some researchers use discriminators of generative adversarial networks (GANs) to 
% prevent physically impossible pose predictions \cite{Kim2021EndtoEndDA}.

\textbf{GCN methods} Since the graph is a natural and suitable tool 
for representing the human body, 
GCN shows great promise in addressing human joint data\cite{Yan2018SpatialTG, Thakkar2018PartbasedGC}.
Cui \etal \cite{Cui2020LearningDR} propose a deep generative model based on graph networks 
and adversarial learning
to tackle the complicated topology problem of human joints.
Dang \etal \cite{Dang2021MSRGCNMR} propose a novel Multi-Scale Residual Graph Convolution Network 
to extract features from fine to coarse scale and then from coarse to fine scale.
Si \etal \cite{Si2019AnAE} integrate the GCN and the LSTM
to solve the communication problem on the temporal dimension. Moreover,
Sofianos \etal \cite{sofianos2021space} use Space-Time-Separable Graph Convolutional Network
to achieve the state-of-the-art. We propose a model to unify the treatment of timing and space.

To construct graph, instead of explicitly defining the graph, 
many researchers have made all joints interconnected
\cite{mao2019learning,mao2020history,sofianos2021space}. 
Nevertheless, the inherent topology of human connections lends itself well to structural comprehension. 
Based on this, we can construct a graph that reflects the natural links between joints.
Partitioning strategies for graphs \cite{Yan2018SpatialTG, Thakkar2018PartbasedGC} 
have also been shown to enhance their effectiveness.
Thakkar \etal \cite{Thakkar2018PartbasedGC} divide the graph into different 
parts and use GCN on different parts respectively. 
Then they use some techniques on common joints of different parts 
to achieve communication between different parts.

\section{Method}
\subsection{Problem Formalization}
\label{sec:sub3.1}
We observe the body pose of a person, represented by the 3D coordinates of its joints, 
for a duration of $T$ frames.
Given $V$ key points (human joints) in each of $T$ frames, the task is to predict the locations of 
these $V$ key points in the next $K$ frames. Each point $v$ at the $k$-th frame is represented by a 
3D vector $\boldsymbol{x}_{v}^{k}$. The historical motion of human poses is represented by the tensor 
$\mathcal{X}_{input}=\left[X_{1}, X_{2} \ldots, X_{T}\right]$, where $X_{i} \in \mathbb{R}^{3 \times V}$. 
The goal is to predict the next $K$ poses $\mathcal{X}_{out}=\left[X_{T+1}, X_{T+2} \ldots, X_{T+K}\right]$. 
Unlike other works, our proposed Multi-Graph Convolutional Network (MGCN) constructs the graph not only 
based on one frame $X_{i}$, but on a series of adjacent frames $\mathcal{X}_{input}$. More details about 
the graph construction will be provided in Section \ref{sec:Multi-Graph Convolution}.

% Given $V$ key points (human joints) in each of $T$ frames,
% we need to predict the locations of $V$ key points in the next $K$ frames.
% % we need to predict the locations of $V$ key points for the future $K$  frames.
% We denote the point $v$ at the $k$-th frame
% by a 3D vector $\boldsymbol{x}_{v}^{k}$. 
% The historical motion of human poses is denoted by the tensor 
% $\mathcal{X}_{input}=\left[X_{1}, X_{2} \ldots, X_{T}\right], 
% X_{i} \in \mathbb{R}^{3 \times V}$. The goal is to predict the next $K$ poses 
% $\mathcal{X}_{out}=\left[X_{T+1}, X_{T+2} \ldots, X_{T+K}\right]$. 
% Unlike other works,  the graph of MGCN we construct is
% based on a series of adjacent frames $\mathcal{X}_{input}$, rather than just one frame.
% % Unlike some other researches,
% % we construct the graph not only on one frame $X_{i}$, 
% % but on a series of consecutive frames $\mathcal{X}_{input}$. 
% We will give more details in \Cref{sec:Multi-Graph Convolution}. 

\subsection{Backgroud of GCN}

Since GCN can easily extract good spatial features, this paper constructs the MGCN model based on it.
%The underlying of MGCN is the GCN, 
%which enables easy extraction of spatial information.
% and generally formalizes as $f\left(\mathcal{X}_{i n} ; A, W\right)$.

Let $\mathcal{H}^{(l)} \in \mathbb{R}^{V \times C^{(l)}}$ be the input of GCN
in the $l$-th layer. Especially, for the first layer, 
the input dimensionality $C^{(1)} =3$ and  $\mathcal{H}^{(1)} = X_{input}$.
The output of the $l$-th layer can be written as
\begin{equation}
  \mathcal{H}^{(l+1)}=\sigma\left(A \mathcal{H}^{(l)} W^{(l)}\right),
  \label{eq:2}
\end{equation}

\noindent
 where $A$ is a constant matrix which is normalized from adjacency matrix,
  $W^{(l)} \in \mathbb{R}^{C^{(l)} \times C^{(l+1)}}$ is the weight 
  and $\sigma$ is the active function.
%  are the trainable 
%  weights of layer $l$ projecting each vertex from $C^{(l)}$  to $C^{(l+1)}$ dimensions. 
% In numerical experiments, we adopt  
% Here our model are implemented by $1 \times 1$ 
%  convolution, and $\sigma$ is an activation function such as ReLU, tanh or PReLU.
 
 \subsection{Multi-Graph Convolution Network}
 \label{sec:Multi-Graph Convolution}
 To construct the graph, we represent the human skeleton as a sparse graph with joints as 
nodes and natural connections between them as edges. 
We define $\mathcal{g}=(\mathcal{v}, \mathcal{e})$ as the human skeleton graph with $V$ nodes, 
and $g\in\mathbb{R}^{V\times V}$ as its spatial adjacency matrix.

If the adjacency matrix is defined according to the traditional method mentioned above, the
receptive field and the amount of information are too small, and the extracted features are incomplete. To avoid this problem, 
we define $g_k$ as a part of $g$. If the length of the shortest path between points i and j 
is equal k, take $g_k$ equal to 1(see \Cref{fig:label2-b}). 
In this paper, all adjacency matrices used in the experiments have the same definition 
as following:

\begin{equation}
    g_{k_{(i,j)}} = 
  \begin{cases}
    \begin{aligned}
      1 &\quad if \quad d\left(i, j\right) = k, \\
      0 &\quad otherwise. \\
    \end{aligned}
  \end{cases}
  \label{eq3.3}
\end{equation}
\begin{equation}
  g = \sum_{k=0}^{D}g_{k} 
\end{equation}
\noindent
where $D$ is a hyperparameter. See details in \Cref{sec:hyperparameter}.

To incorporate the interactions of body joints across all observed frames,
 we introduce the graph with $T*V$ nodes denoted by $\mathcal{G}=(\mathcal{V}, \mathcal{E})$, 
 with its adjacency matrix being $G\in\mathbb{R}^{VT\times VT}$. As shown in Figure \ref{fig:label2-a}, 
 $G$ is constructed by replicating $g$ along the diagonal and connecting the same joints across different 
 frames within span of \textbf{L}(\Cref{sec:hyperparameter}). 
 The connections of $\mathcal{E}$ are defined as follows. Within each frame, we add edges between 
 connected joints in $\mathcal{g}$; across different frames, we add edges between joints that correspond 
 to the same body part. The connections of $\mathcal{E}$ are formulated by

\begin{equation}
  G\left(\boldsymbol{x}_{v1}^{k1},\boldsymbol{x}_{v2}^{k2}\right) = 
  \begin{cases}
    \begin{aligned}
      1 \quad if  \left(\boldsymbol{x}_{v1}, \boldsymbol{x}_{v2}\right) \in \mathcal{e}, \\
      0 \quad if  \left(\boldsymbol{x}_{v1}, \boldsymbol{x}_{v2}\right) \notin \mathcal{e}. \\
    \end{aligned}
  \end{cases}
  \label{eq:1}
\end{equation}
\noindent
where $\boldsymbol{x}_{v1}^{k1}$ indicates the $v1$ node in frame $k1$, same as $\boldsymbol{x}_{v2}^{k2}$.
And $\vert k1-k2 \vert \leq L$.

The multi-graph construction enables our model to capture temporal dependencies and 
 spatial interactions among joints.
% \noindent
% For the joints in the same frame, 
% we construct the graph along the kinematic. For the joints in different frames, 
% we connect the joints which are naturally connected in one frame.
% The whole MGCN can be modeled as 
%  \begin{equation}
%   \begin{aligned}
%   \hat{\mathcal{X}}_{input} = MGCN(\mathcal{X}_{input}), \\ \hat{\mathcal{X}}_{input},
%   \mathcal{X}_{input} \in \mathbb{R}^{ T \times V \times 3}.
%   \label{eq:4}
%   \end{aligned} 
% \end{equation}
It is worth noting that the dimensions of input and output are same:
\begin{equation}
  \label{eq:MGCN}
  MGCN(\mathcal{X}_{input}, G):\mathbb{R}^{ T \times V \times 3} \rightarrow \mathbb{R}^{ T \times V \times 3}.
\end{equation}
% $\mathbb{R}^{ T \times V \times 3} \rightarrow \mathbb{R}^{ T \times V \times 3}$.
The $l$-th layer of 
the multi-graph convolution layer is 
\begin{equation}
  \begin{aligned}
   \mathcal{H}^{(l+1)}= \sigma\left(G \mathcal{H}^{(l)} W^{(l)}\right)=\sigma\left(\sum_{k=0}^{D}G_{k}
   \mathcal{H}^{(l)} W^{(l)}_{k}\right)
   \label{eq:3}
  \end{aligned}
\end{equation}
% \begin{equation}
%   G = \sum_{k=0}^{D}G_{k} 
% \end{equation}
where $G_{k}$ is an integrated matrix which consists of $g_{k}$.

\subsection{Hyperparameter L \& D} 
\label{sec:hyperparameter}
% To avoid joint interactions over the entire time span, 
We incorporate a hyperparameter \textbf{L} to limit the range of joint connections. 
Specifically, in our experiments, we only connect the t-th frame to 
frames between the $(t-L)$-th and $(t+L)$-th frames (if they exist). This parameter 
controls the spatial receptive field and aids in extracting robust information for different tasks. 

\begin{figure*}[!htbp]
  \centering
  \begin{subfigure}{0.68\linewidth}
    \includegraphics[width=1\linewidth]{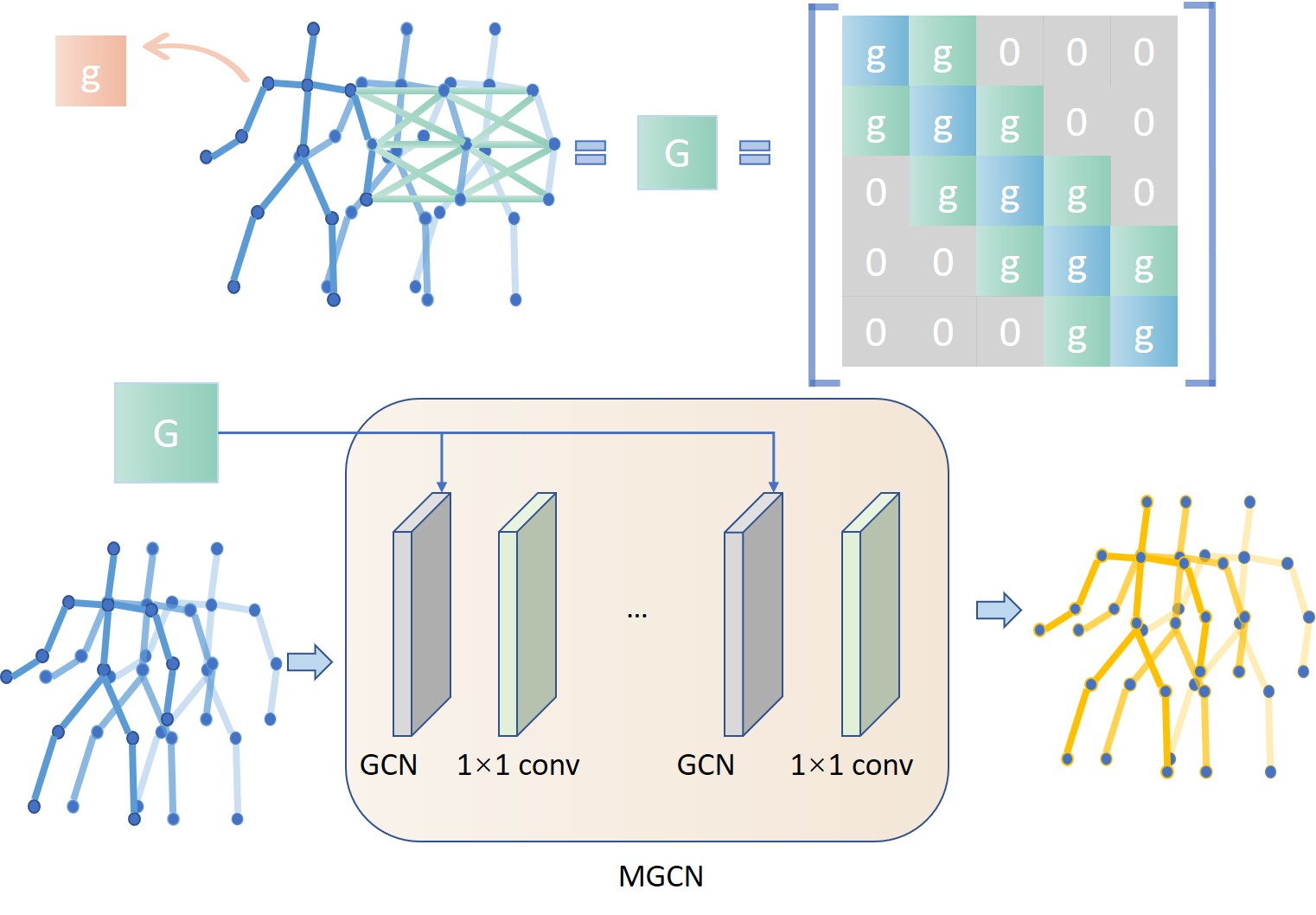}
    \caption{}
    \label{fig:label2-a}
  \end{subfigure}
  \hfill
  \begin{subfigure}{0.28\linewidth}
    \includegraphics[width=1\linewidth]{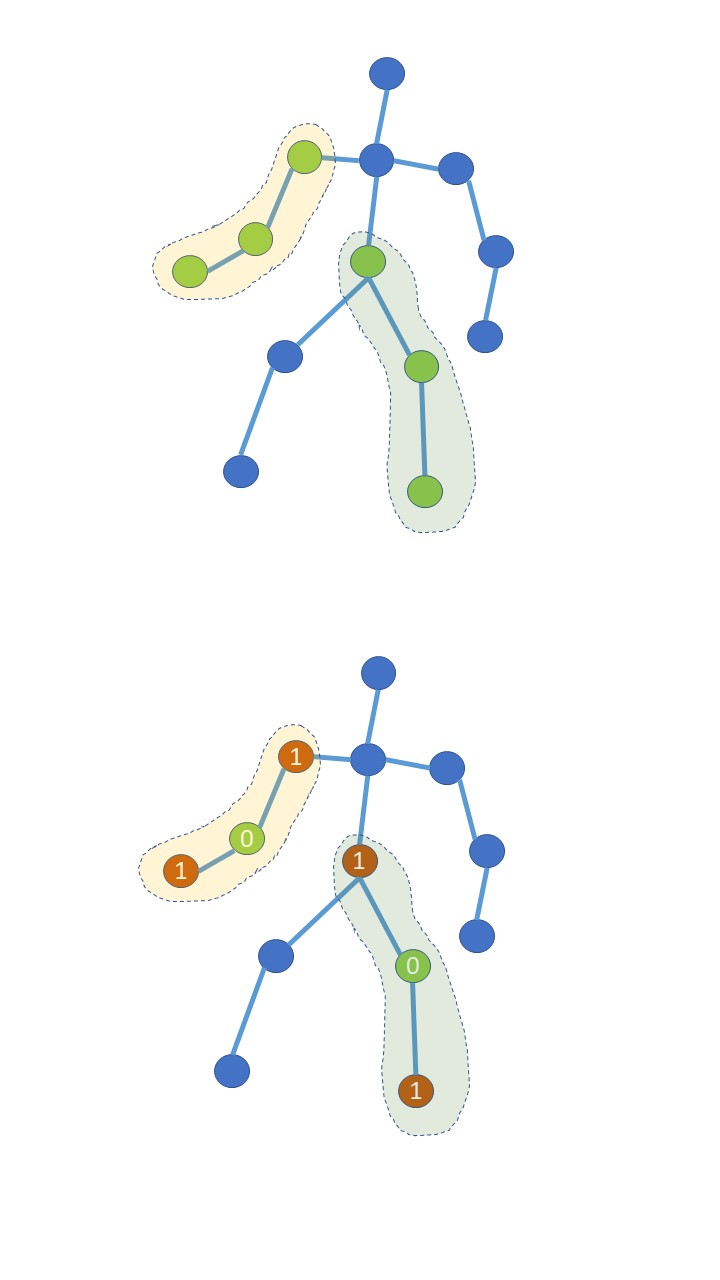}
    \caption{}
    \label{fig:label2-b}
  \end{subfigure}
  \caption{Overview of the MGCN model and Distance partition. 
  For convenience, we assume that the human body has 13 joints. 
  (a) illustrates the model MGCN. The graph G for MGCN  consists of $T \times T$ g,
  where $T$ is the number of input frames. 
  The blue parts show the connections inner frames; green parts show the connections inter frames; 
  gray parts show the frames beyond time receptive field $L$.
  Here 
  we illustrate the state when $T$ = 5, $L$ = 1.
  For every frame $t$, we merely connect from
  the frame $t-1$ to $t+1$ (if they exist).
  Given joints and G as input, MGCN outputs the prediction.
  (b) is the illustrations with or without  Distance partition.}
  \label{fig:label2}
\end{figure*}

Additionally, we utilize another hyperparameter \textbf{D} to restrict the maximum distance that a 
joint can connect to in one frame. 
We follow the Distance partition\cite{Yan2018SpatialTG} to divide the adjacency matrix $g$ according to different hops
% the length of two nodes
(see \Cref{fig:label2-b}).
Our experiments demonstrate that setting a 
large \textbf{L} and \textbf{D} is not 
beneficial for long-term forecasting, as shown in \Cref{tab:experiment}.

\subsection{MGCN with sequence-aware attention} %{Attentioned-MGCN}
\label{sec:3.5}

Many recent works on pose forecasting have used self-attention mechanisms to capture the 
relationship between frames \cite{mao2020history, Cai2020LearningPJ} and/or the relationship 
of joints \cite{Cai2020LearningPJ}. We adopt the Multi-Graph Convolutional Network 
(MGCN) to produce the Query (Q), Key (K), and Value (V) representations. 
Due to the sequential nature of the data, 
we use masked attention which we refer it as sequence-aware attention.
In contrast to using position encoding, which yielded 
suboptimal results, we propose two novel strategies for sequence-aware attention: Pseudo-autoregressive 
and Anchor.

\noindent
\textbf{Pseudo-autoregressive strategy:} We directly set
the $S$ matrix (see \Cref{fig:label1}) as a
lower triangular matrix filled with 1. 
To achieve better results, we predict the coordinate offset of the original position instead 
of predicting the position directly.
S * V achieves offset accumulation in the time dimension.
%  that the value in frame $T+i$ is obtained by adding the 
% predicted offset to the previous value $X_{T+i-1}$. 
The outputs are defined as follows:
\begin{equation}
  % \label{eq:6}
  offset = MGCN_V(X_{input},G)
  % X_{T+i} =\sum_{k=1}^{i} offset_{k}+ X_{T}. 
\end{equation}
\begin{equation}
  \label{eq:5}
  X_{T+i} =\sum_{k=1}^{i} offset_{k}+ X_{T},\quad offset_{k} \in \mathbb{R}^{ V \times 3}.
\end{equation}

\begin{figure}[!hbtp]
  \centering
     \includegraphics[width=0.8\linewidth]{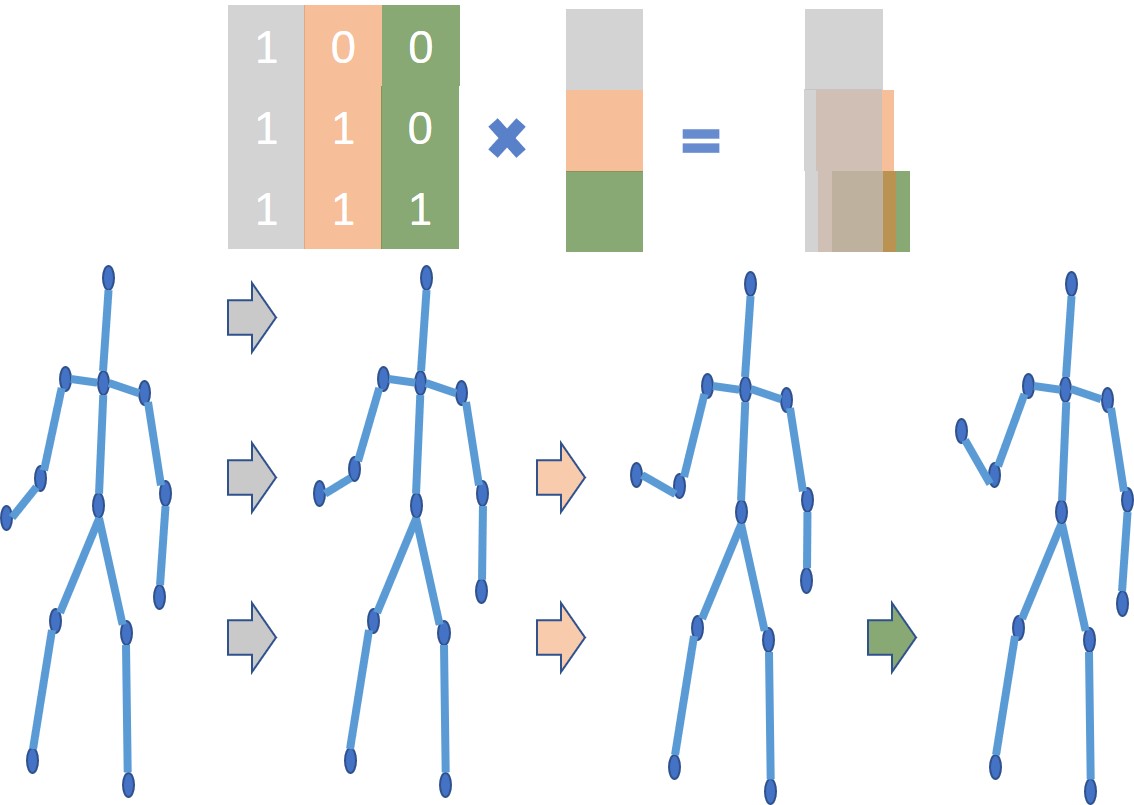}  
     \caption{The illustration of the Pseudo-autoregressive strategy of sequence-aware attention. The output of time $t$ can be considered as 
     a recursive addition of the previous offsets, so the model can have a global reception of movement.
     }
  \label{fig:armodel}
\end{figure}

The process of predicting future frames based 
on previous data is illustrated in Figure \ref{fig:armodel}. This approach not only 
avoids the need to compute MGCN-Q and MGCN-K, but also achieves the best performance 
in short-term prediction (as discussed in more detail in Section \ref{sec:ab}).

\begin{figure}
    \centering
       \includegraphics[width=0.8\linewidth]{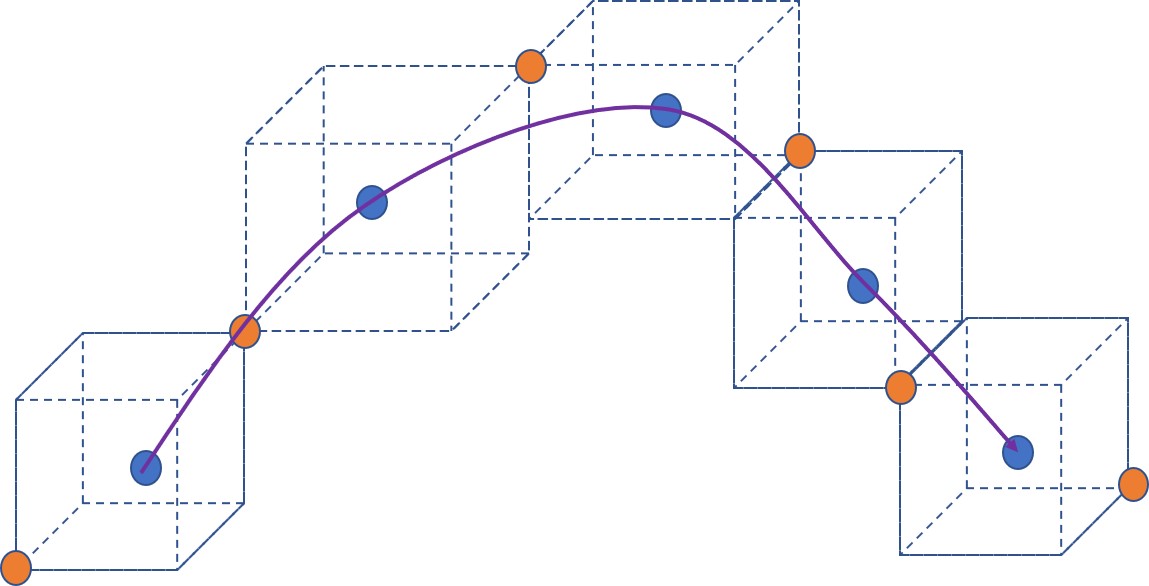}    
       \caption{The illustration of the Anchor strategy. This strategy uses anchors to restrict the reachable space of the 
       future joints. The yellow points on the corner of the cuboid are anchors, so the cuboid formed by two yellow anchors 
       is the space where the future joint will be. 
       We only use the last two anchors to form the reachable cuboid for the convenience of description.
       }
    \label{fig:anchormodel}
\end{figure}

\noindent
\textbf{Anchor (Possible Space):} We introduce Anchor to add
spatial constraints into our model.
Firstly S matrix is generated by MGCN-Q and MGCN-K through masked softmax, 
and a set of anchors are produced by MGCN-V.
The anchors work as a set of basis for Possible Space of future joints. 
Specially, it can be constructed as a cuboid along axis when there are only two anchors (\Cref{fig:anchormodel}).
Specifically, we use masked softmax 
on each of the three spatial dimensions (x, y, z), resulting in a reachable space that is a cuboid.
The prediction for each joint is then a convex combination of these anchor points. 
The output in the frame $T+i$ is given by the following equation:
\begin{equation}
  \label{eq:6}
  \begin{aligned}
    X_{T+i}^{d} &= \sum_{k=1}^{i}\lambda_{ik}^{d}anchor_{k}^{d}, \quad anchor_{k} \in \mathbb{R}^{ V \times 3}
  \end{aligned}
\end{equation}
\begin{equation}
  anchor = MGCN_V(X_{input},G),
\end{equation}
where $\sum_{k=1}^{i}\lambda_{ik}^{d}=1$, $d \in {1, 2, 3}$.
% Here, we determine a possible space for 
% the future joints to reach and predict the most likely location within that space.
\Cref{fig:anchormodel} illustrates the process of predicting by 2 anchors.
In our model, we use the 10 anchors to have a better understanding of the global movement.”

% where $\sum_{k=1}^{i}\lambda_{ik}^{d}=1$, d = 1, 2, 3.
% The $d$ means the different dimensions (x,y,z). We first determine a possible space the future joints can reach 
% and find the most likely location. We 
% use the $i$-th input as the base position to calculate the anchor. 
% \Cref{fig:anchormodel} illustrates the state when we only use the last 
% 2 anchors to predict. It is slightly worse than the best on short-term prediction. 
% In our model, we use the last 10 anchors before 
% so that we can have a better understanding of global movement. 

\subsection{TCN \& Refine MGCN}
% Next, a Temporal Convolutional Network (TCN) module is applied to capture the temporal dynamics. 
% Finally, a refined MGCN is used to improve the performance of the frame output.

% After incorporating an improved attention mechanism, 
% The prediction results are obtained for the same time span as the input data. 
The time dimension of the result after attention is same as the input (\Cref{eq:MGCN}).
To fix the temporal mismatch between the output and target,
we utilize Temporal Convolutional 
Networks to align the time dimension following \cite{sofianos2021space}. 
Specifically, the 1*1 conv are applied in the temporal dimension.

Finally, a refined MGCN is used to improve the performance of the frame output.
% to align the dimension. 
% TCNs are a type of decoder commonly used in sequence 
% modeling, and they are preferred for their simplicity and robustness compared to other modeling options 
% like LSTM ~\cite{Hochreiter1997LongSM}, GRU ~\cite{Cho2014LearningPR}, and Transformer Networks 
% \cite{Giuliari2021TransformerNF}.

% \subsection{Refine MGCN}

\begin{table}[!htbp]
  \centering
  \begin{tabular}{@{}lccc@{}}
    \toprule
    msec & 80& 400 &1000\\
    \midrule
    DCT-RNN-GCN\cite{mao2020history}& 10.4   &58.3&112.1\\
    BGPIG\cite{Ma2022ProgressivelyGB}& 10.3  &58.5& -   \\
    STS-GCN\cite{sofianos2021space} & 10.1   &38.3&75.6 \\
    \midrule
    SAA-10-MGCN L2D3      & 8.9            & 35.1         & \textbf{72.9} \\
    SAA-10-MGCN L6D4      & \textbf{8.6} & \textbf{34.4} & 83.6 \\
  %   sequence-aware attention_L25_D4      & $\mathbf{ }$  &$\mathbf{}$&$\mathbf{89.3}$ \\
  %   sequence-aware attention_L3_D4      & $\mathbf{ }$  &$\mathbf{}$&$\mathbf{75.8}$ \\
  %   sequence-aware attention_L3_D4_[16,16,16]      & $\mathbf{ }$  &$\mathbf{}$&$\mathbf{73.88}$ \\
  %   sequence-aware attention_L2_D4_[16,16,16]      & $\mathbf{ }$  &$\mathbf{}$&$\mathbf{72.8}$ \\
  %   sequence-aware attention_L2_D2_[16,16,16]      & $\mathbf{ }$  &$\mathbf{}$&$\mathbf{76.4}$ \\
  %   sequence-aware attention_L1_D4_[16,16,16]      & $\mathbf{ }$  &$\mathbf{}$&$\mathbf{75.9}$ \\
  %   sequence-aware attention_L1_D1_[16,16,16]      & $\mathbf{ }$  &$\mathbf{}$&$\mathbf{73.4}$ \\
    %sequence-aware attention_L10_D10      & $\mathbf{ }$  &$\mathbf{}$&$\mathbf{89.3}$ \\
    \bottomrule
  \end{tabular}
  \caption{MPJPE error in mm for prediction of 3D joint positions on Human3.6M. All are given 10 frames to 
  predict the 2, 10, 25 frames(80, 400, 1000 msec) in the future respectively. 
  \textbf{SAA-10-MGCN} means the model using sequence-aware attention and 10 anchors before.
  \textbf{L2D3} indicates we set L = 2, D = 3 in our MGCN.
  Our model outperforms the state-of-the-art on three scales. 
  We change the temporal receptive field L and graph max hop distance D for different tasks.}
  \label{tab:experiment}
\end{table}

\begin{table*}
  \centering
  \resizebox{\linewidth}{!}{
  
  \begin{tabular}{@{}lcccccccccccccccc@{}}
    \cline{2-17}
                  & \multicolumn{4}{|c|}{Walking} & \multicolumn{4}{c|}{Eating}
                  & \multicolumn{4}{c|}{Smoking} & \multicolumn{4}{c|}{Discussion }  \\ \midrule
    msec         & 80& 160 &320& 400  & 80& 160 &320& 400 & 80& 160 &320& 400 & 80& 160 &320& 400           \\
    \midrule
    DCT-RNN-GCN\cite{mao2020history}& 10.0   &19.5&34.2&39.8&6.4&14.0&28.7&36.2&7.0&14.9&29.9&36.4&10.2&23.4&52.1&65.4\\
    STS-GCN\cite{sofianos2021space} & 10.7   &16.9&29.1&32.9&6.8&11.3&22.6&25.4&7.2&11.6&22.3&25.8&9.8 &16.8&33.4&40.2\\
    BGPIG\cite{Ma2022ProgressivelyGB}&10.2 &19.8 &34.5 &40.3 &7.0 &15.1 &30.6& 38.1& 6.6& 14.1& 28.2& 34.7& 10.0& 23.8& 53.6& 66.7\\
    \midrule
    Ours (L6D4) &\textbf{9.3}&	\textbf{14.0}&	\textbf{24.7}&	\textbf{28.6}&\textbf{	5.7}&	\textbf{10.2}&	\textbf{18.9}&	\textbf{22.3}&	
    \textbf{5.8}&	\textbf{10.9}&	\textbf{20.0}&	\textbf{22.9}&	\textbf{8.0}&	\textbf{14.6}&	\textbf{29.8}&	\textbf{35.9}
    \\ \\
    \cline{2-17}
                    & \multicolumn{4}{|c|}{Directions } & \multicolumn{4}{c|}{Greeting }
                    & \multicolumn{4}{c|}{Phoning} & \multicolumn{4}{c|}{Posing }  \\ \midrule
      msec         & 80& 160 &320& 400  & 80& 160 &320& 400 & 80& 160 &320& 400 & 80& 160 &320& 400        \\
      \midrule
      DCT-RNN-GCN\cite{mao2020history} & 7.4&18.5&44.5&56.5&13.7&30.1&63.8&78.1&8.6&18.3&39.0&49.2&10.2&24.2&58.2&75.8\\
      STS-GCN\cite{sofianos2021space}  & 7.4&13.5&29.2&34.7&12.4&21.8&42.1&49.2&8.2&13.7&26.9&30.9&9.9&18.0&38.2&45.6 \\
      BGPIG\cite{Ma2022ProgressivelyGB}&7.2& 17.6& 40.9& 51.5& 15.2& 34.1& 71.6& 87.1& 8.3& 18.3& 38.7& 48.4& 10.7& 25.7& 60.0& 76.6\\
      \midrule
      Ours (L6D4) &\textbf{5.9}&	\textbf{11.5}&	\textbf{24.5}&	\textbf{30.6}&	\textbf{10.8}&	\textbf{18.3}&	\textbf{35.9}&
      	\textbf{44.5}&	\textbf{6.8}&	\textbf{12.5}&	\textbf{23.1}&	\textbf{27.7}&	\textbf{7.7}&	\textbf{15.5}&	\textbf{32.4}&	\textbf{40.9}
      \\ \\
      \cline{2-17}
                    & \multicolumn{4}{|c|}{Purchases} & \multicolumn{4}{c|}{Sitting }
                    & \multicolumn{4}{c|}{Sitting Down} & \multicolumn{4}{c|}{Taking Photo}  \\ \midrule
      msec         & 80& 160 &320& 400  & 80& 160 &320& 400 & 80& 160 &320& 400 & 80& 160 &320& 400        \\
      \midrule
      DCT-RNN-GCN\cite{mao2020history}& 13.0&29.2&60.4&73.9&9.3&20.1&44.3&56.0&14.9&30.7&59.1&72.0&8.3&18.4&40.7&51.5\\
      STS-GCN\cite{sofianos2021space} & 11.9&21.3&42.0&48.7&9.1&15.1&29.9&35.0&14.4&23.7&41.9&47.9&8.2&14.2&29.7&33.6\\
      BGPIG\cite{Ma2022ProgressivelyGB}&12.5& 28.7& 60.1& 73.3& 8.8& 19.2& 42.4& 53.8& 13.9& 27.9& 57.4& 71.5& 8.4& 18.9& 42.0& 53.3\\
      \midrule
      Ours (L6D4) &\textbf{10.2}&	\textbf{18.6}&	\textbf{37.1}&	\textbf{44.2}&	\textbf{7.6}&	\textbf{14.1}&	\textbf{26.6}&	\textbf{31.8}
      &  \textbf{14.0}&	\textbf{21.7}&	\textbf{37.8}&	\textbf{45.5}&	\textbf{7.0}&	\textbf{13.3}&	\textbf{25.4}&	\textbf{30.9}
      \\\\
      \cline{2-17}
                    & \multicolumn{4}{|c|}{Waiting } & \multicolumn{4}{c|}{Walking Dog }
                    & \multicolumn{4}{c|}{Walking Together} & \multicolumn{4}{c|}{Average}  \\ \midrule
      msec         & 80& 160 &320& 400  & 80& 160 &320& 400 & 80& 160 &320& 400 & 80& 160 &320& 400        \\
      \midrule
      DCT-RNN-GCN\cite{mao2020history}& 8.7&19.2&43.4&54.9&20.1&40.3&73.3&86.3&8.9&18.4& 35.1& 41.9& 10.4& 22.6& 47.1& 58.3\\
      STS-GCN\cite{sofianos2021space} & 8.6&14.7&29.6&35.2&17.6&29.4&52.6& 59.6&8.6&14.3&26.5&30.5&10.1&17.1&33.1&38.3\\
      BGPIG\cite{Ma2022ProgressivelyGB}&8.9& 20.1& 43.6& 54.3& 18.8& 39.3& 73.7& 86.4& 8.7& 18.6& 34.4& 41.0& 10.3& 22.7& 47.4& 58.5\\
      \midrule
    Ours (L6D4) &\textbf{6.8}&	\textbf{12.4}&	\textbf{24.9}&	\textbf{30.3}&	\textbf{15.6}&	\textbf{25.5}&	\textbf{45.5}&	\textbf{54.3}
    &	\textbf{7.3}&	\textbf{12.4}&	\textbf{22.4}&	\textbf{26.0}&	\textbf{8.6}&	\textbf{15.0}&	\textbf{28.6}&	\textbf{34.4}
    \end{tabular}}
  \caption{ MPJPE error in mm for short-term prediction of 3D joint positions on Human3.6M. Our model (L6D4) outperforms the
  state-of-the-art by a large margin. We use 10 frames (400 msec) as input and predict 2-10 frames(80-400 msec).}
\label{tab:1}
\end{table*}

\subsection{Training}

In the MGCN architecture of end-to-end supervised training, we adopt the loss function
based on Mean Per Joint Position Error (MPJPE) ~\cite{Ionescu2014Human36MLS,mao2019learning}
% Supervision is provided by the losses
% that measure error wrt ground truth in terms of 
% Mean Per Joint Position Error (MPJPE) ~\cite{Ionescu2014Human36MLS,mao2019learning}.  
% The loss based on MPJPE
% is:
\begin{equation}
   {\cal L}_{M P J P E}=\frac{1}{V \times K} \sum_{k=T}^{T+K} \sum_{v=1}^{V}\left\|\hat{\boldsymbol{x}}_{v}^{k}-\boldsymbol{x}_{v}^{k}\right\|_{2}
   \label{eq:7},
\end{equation}
where $\hat{\boldsymbol{x}}_{v}^{k} \in \mathbb{R}^{3}$ is the corresponding ground truth
and $\boldsymbol{x}_{v}^{k} \in \mathbb{R}^{3}$ denotes the predicted coordinates of the joint $v$ 
in the $k$-th frame.

\section{Experimental evaluation}
We experimentally evaluate the proposed model against 
the state-of-the-art on the recent, large-scale and challenging benchmark, Human3.6M, AMASS and 3DPW. 

% \begin{figure*}
%   \begin{subfigure}{0.7\linewidth}
%     \includegraphics[width=0.8\linewidth]{eat_walk.jpg}  
%      \caption{Sample long-term predictions (25 frames, 1 sec) for the actions of Eating and Walking.
%      Purple/green limbs are the left/right sides of the body. Gray/black pictorials
%     indicate the observed ground-truth (GT) skeletons.}
%   \label{fig:visualization}
%   \end{subfigure}
%   \hfill
%   \begin{subfigure}{0.3\linewidth}
%     \begin{subfigure}{1\linewidth}%[position][height][inner pos]{width}
%       \includegraphics[width=1\linewidth]{line_fig2.jpg}  
%       \caption{}
%       \end{subfigure}
%       \begin{subfigure}{1\linewidth}%[position][height][inner pos]{width}
%       \includegraphics[width=1\linewidth]{line_fig4.jpg}  
%       \caption{}
%       \end{subfigure} 
%         \caption{(a) The influence of hyperparameters L and D. 
%         The line plot shows that the best performance of the model is obtained when L=4 and D=6. 
%         (b)The advantage of our
%         method is most significant for the action of “walking dog”.}
%       \label{fig:line_fig2}
%   \end{subfigure}
% \end{figure*}

\begin{figure}[!hbtp]
  \centering
  \begin{subfigure}{1\linewidth}%[position][height][inner pos]{width}
  \includegraphics[width=1\linewidth]{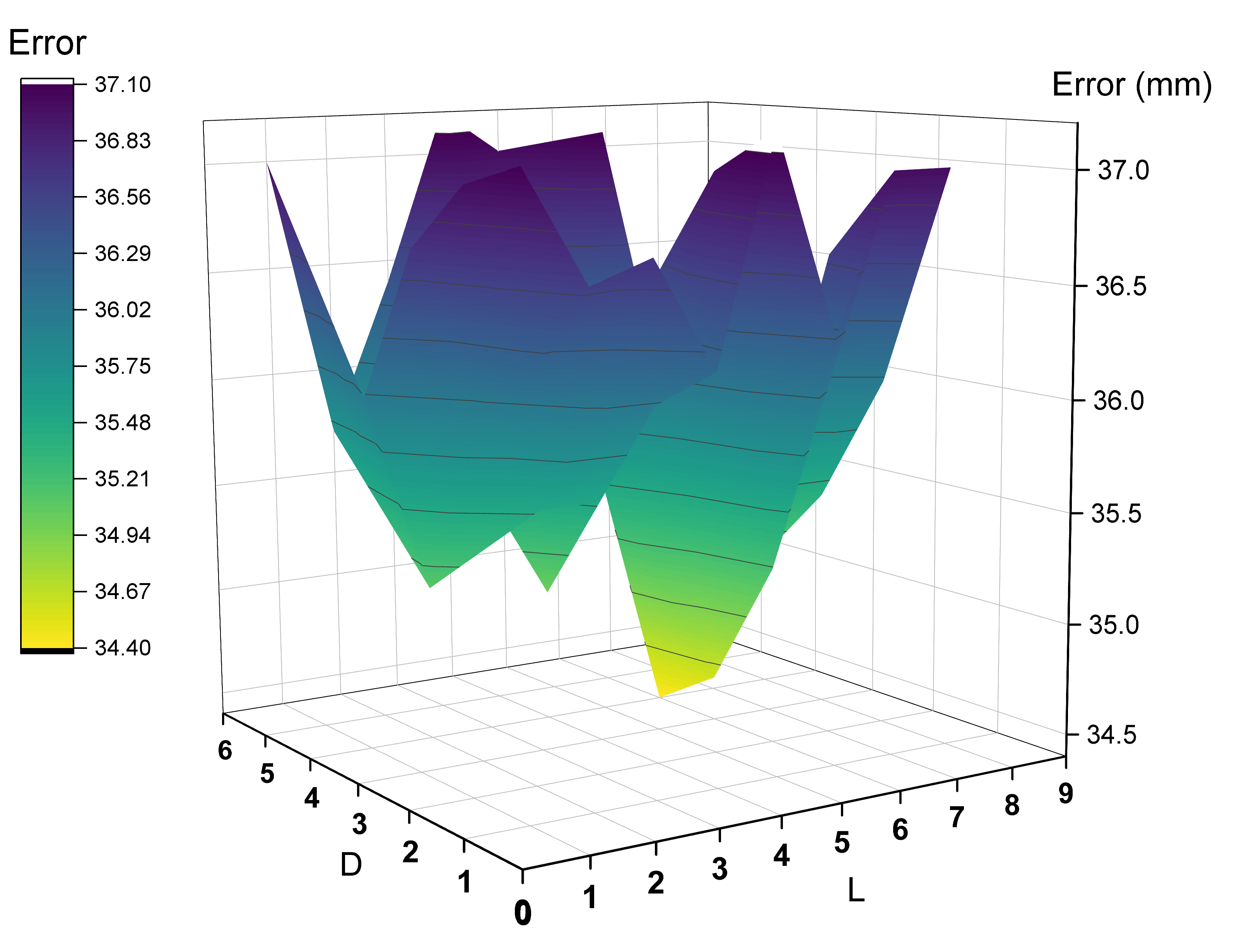}  
  \caption{}
  \end{subfigure}
  \begin{subfigure}{1\linewidth}%[position][height][inner pos]{width}
  \includegraphics[width=1\linewidth]{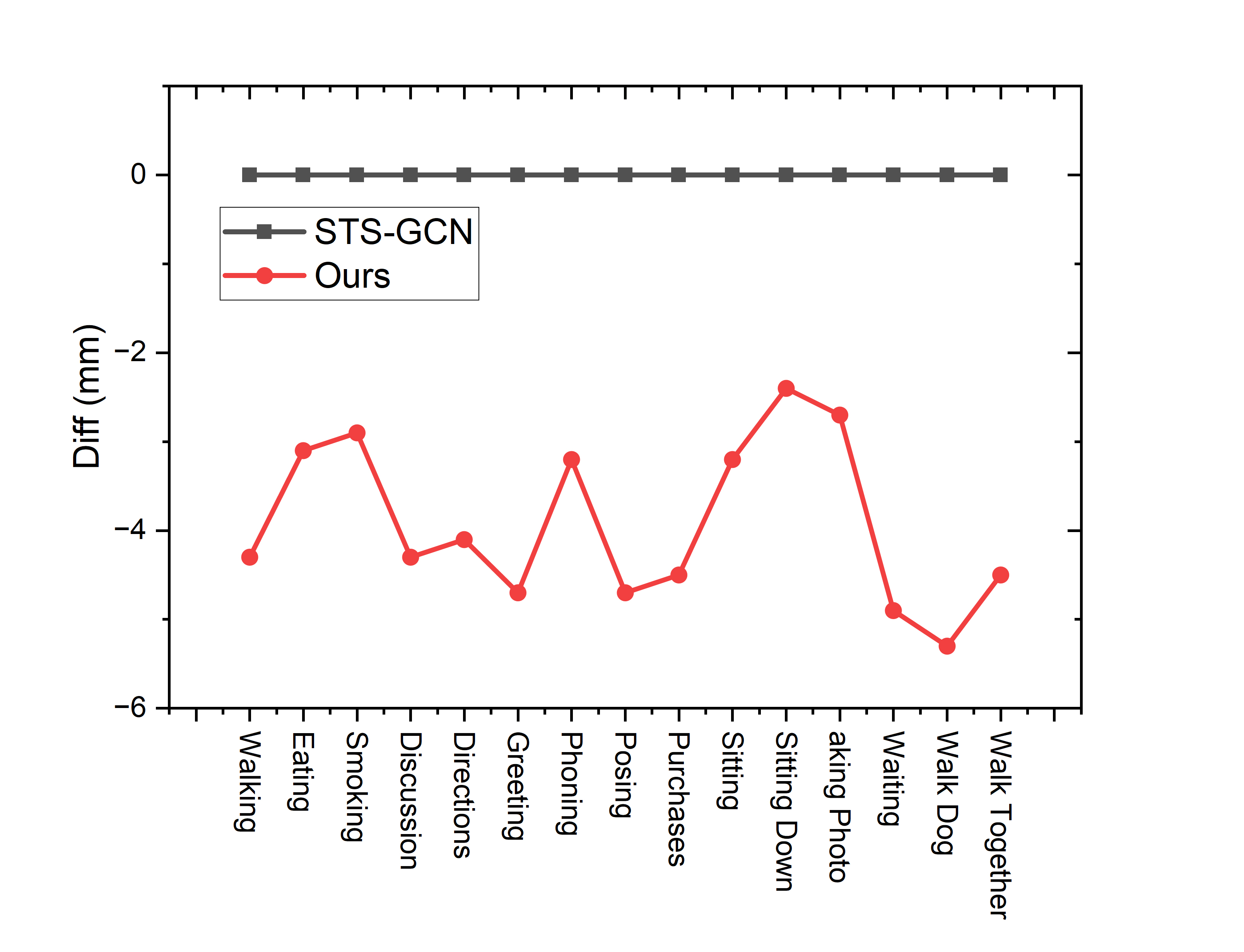}  
  \caption{}
  \end{subfigure} 
    \caption{(a) The influence of hyperparameters L and D. 
    The plot shows that the best performance of the model is obtained when L=4 and D=6 on 400ms task. 
    (b)The advantage of our
    method is most significant for the action of “walking dog”.}
  \label{fig:line_fig2}
\end{figure}

\subsection{Datasets and metrics}
\textbf{Human3.6M.}\cite{Ionescu2014Human36MLS} This dataset is widespread for human pose
forecasting, which has 3.6 million 3D human poses and the
corresponding images. It consists of 7
actors performing 15 different actions (e.g. Walking, Eating, Phoning). The actors are represented as skeletons of
32 joints. The orientations of joints are represented as exponential maps, from which the 3D coordinates may be computed 
\cite{Taylor2006ModelingHM,Fragkiadaki2015RecurrentNM}.
For each pose, we consider 22 joints out of the provided 32 for estimating MPJPE\cite{sofianos2021space}.
Following the current literature \cite{mao2019learning,mao2020history,Martinez2017OnHM}, 
we use the subject 11 (S11) for validation, the subject 5 (S5) for testing,
and all the rest of the subjects for training.

\textbf{AMASS}\cite{Mahmood2019AMASSAO} The Archive of Motion Capture as Surface
Shapes (AMASS) dataset has been recently proposed, to
gather 18 existing mocap datasets. 
We select 10 from those and take 5 for training, 4 for validation
and 1 (BMLrub) as the test set. Then we use the SMPL \cite{Loper2015SMPLAS}
parameterization to derive a representation of human pose
based on a shape vector
and its joints rotation angles. We obtain human poses in 3D
by applying forward kinematics. Overall, AMASS consists
of 40 human-subjects that perform the action of walking.
Each human pose is represented by 52 joints, including 22
body joints and 30 hand joints. Here we consider for forecasting the 
body joints only and discard from those 4 static
ones, leading to an 18-joint human pose. 

\textbf{3DPW}\cite{Marcard2018RecoveringA3} The 3D Pose in the Wild dataset consists
of video sequences acquired by a moving phone camera.
3DPW includes indoor and outdoor actions. Overall, it contains 51,000 frames captured at 
30Hz, divided into 60 video
sequences. We use this dataset to test generalization of the
models trained on AMASS.

\textbf{Metrics} Following the benchmark protocols, we adopt the 
MPJPE error metrics. It quantifies the error of the 3D coordinate predictions in mm. 
We only adopted the 3D coordinate due to the inherent ambiguity of angle 
representation \cite{Cai2020LearningPJ,Akhter2015PoseconditionedJA}.

% Due to the angle representation suffers from an inherent
% ambiguity, and MPJPE is more effective \cite{Cai2020LearningPJ,Akhter2015PoseconditionedJA}. 
% Hence, we adopted the 3D coordinate here.

\textbf{Implementation details} The MGCN is a 
seq2seq model which takes a series of joints from 
consecutive frames as inputs, 
and its outputs represent a series of vectors 
in the future consecutive frames. 
Following \cite{sofianos2021space},
we stack MGCN 
modules and only channel C changes: 
from 3 (the input dimension: x, y, z) to 64, then 32, 64,
and finally 3. 
MGCN for Q and K change the channel by \{3,64,32,16,16,3\}. 
The masked softmax is adopted to implement sequence-aware 
attention. 
We train the model using the Adam optimizer on
one NVIDIA RTX 2080Ti GPU with the minibatch size for
GPU set as 256. The number of total epochs was fixed at
50. The learning rate is initially set to 0.1 and decays only
at positions \{20, 35, 45\}.
%%%the learning rate is initially set to 0.1 and decayed by a factor of 0.1 for 50 epochs, which milestones 
%%%are \{20, 35, 45\}. (123)
% the learning rate is initially set to 0.1 and decays only at positions \{20, 35, 45\} for a total of 50 epochs.
\begin{table*}
  \centering
  \resizebox{\linewidth}{!}{
  
  \begin{tabular}{@{}lcccccccccccccccc@{}}
    \cline{2-17}
                  & \multicolumn{4}{|c|}{Walking} & \multicolumn{4}{c|}{Eating}
                  & \multicolumn{4}{c|}{Smoking} & \multicolumn{4}{c|}{Discussion }  \\ \midrule
    msec         & 560& 720 &880& 1000  & 560& 720 &880& 1000& 560& 720 &880& 1000 & 560& 720 &880& 1000           \\
    \midrule
    DCT-RNN-GCN\cite{mao2020history}& 47.4& 52.1& 55.5& 58.1 &50.0& 61.4&70.6&75.5&47.5 &56.6& 64.4& 69.5&86.6& 102.2& 113.2 &119.8\\
    STS-GCN\cite{sofianos2021space} & 40.6 &45.0 &48.0 &51.8&33.9 &40.2 &46.2 &52.4&33.6 &39.6 &45.4 &50.0&53.4 &63.6 &72.3 &78.8\\
    \midrule
    Ours (L2D3) &\textbf{36.0}&	\textbf{41.6}&	\textbf{44.2}&	\textbf{47.5}&	\textbf{31.5}&	\textbf{38.6}&	\textbf{44.0}&	\textbf{48.8}&
    	\textbf{32.8}&	\textbf{40.1}&	\textbf{43.9}&	\textbf{49.1}&	\textbf{50.8}&	\textbf{62.2}&	\textbf{70.2}&	\textbf{76.8}
    \\\\
    \cline{2-17}
                    & \multicolumn{4}{|c|}{Directions } & \multicolumn{4}{c|}{Greeting }
                    & \multicolumn{4}{c|}{Phoning} & \multicolumn{4}{c|}{Posing }  \\ \midrule
      msec         & 560& 720 &880& 1000  & 560& 720 &880& 1000& 560& 720 &880& 1000 & 560& 720 &880& 1000         \\
      \midrule
      DCT-RNN-GCN\cite{mao2020history} & 73.9 &88.2 &100.1 &106.5&101.9 &118.4 &132.7 &138.8&67.4 &82.9 &96.5 &105.0&107.6 &136.8 &161.4 &178.2\\
      STS-GCN\cite{sofianos2021space}  & 47.6 &56.5 &64.5 &71.0&64.8 &76.3 &85.5 &91.6&41.8 &51.1 &59.3 &66.1&64.3 &79.3 &94.5 &106.4 \\
      \midrule
      Ours (L2D3) &\textbf{44.1}&	\textbf{54.7}&	\textbf{63.6}&	\textbf{69.9}&	\textbf{60.8}&	\textbf{73.2}&	\textbf{83.9}&	\textbf{89.3}&
      	\textbf{40.0}&	\textbf{50.1}&	\textbf{56.6}&	\textbf{63.9}&	\textbf{60.2}&	\textbf{75.8}&	\textbf{89.8}&	\textbf{100.3}
      \\\\
      \cline{2-17}
                    & \multicolumn{4}{|c|}{Purchases} & \multicolumn{4}{c|}{Sitting }
                    & \multicolumn{4}{c|}{Sitting Down} & \multicolumn{4}{c|}{Taking Photo}  \\ \midrule
      msec        & 560& 720 &880& 1000  & 560& 720 &880& 1000& 560& 720 &880& 1000 & 560& 720 &880& 1000       \\
      \midrule
      DCT-RNN-GCN\cite{mao2020history}& 95.6 &110.9 &125.0 &134.2&76.4 &93.1 &107.0 &115.9&97.0 &116.1 &132.1 &143.6&72.1 &90.1 &105.5 &115.9\\
      STS-GCN\cite{sofianos2021space} & 63.7 &74.9 &86.2 &93.5&47.7 &57.0 &67.4 &75.2&63.3 &73.9 &86.2 &94.3&47.0 &57.4 &67.2 &76.9\\
      \midrule
      Ours (L2D3) &\textbf{60.6}&	\textbf{74.7}&	\textbf{85.1}&	\textbf{91.2}&	\textbf{46.0}&	\textbf{57.9}&	\textbf{65.4}&	\textbf{73.0}&
       \textbf{61.7}&	\textbf{74.9}&	\textbf{85.5}&	\textbf{93.2}&	\textbf{45.7}&	\textbf{56.3}&	\textbf{65.8}&	\textbf{73.2}
      \\\\
      \cline{2-17}
                    & \multicolumn{4}{|c|}{Waiting } & \multicolumn{4}{c|}{Walking Dog }
                    & \multicolumn{4}{c|}{Walking Together} & \multicolumn{4}{c|}{Average}  \\ \midrule
      msec         & 560& 720 &880& 1000  & 560& 720 &880& 1000& 560& 720 &880& 1000 & 560& 720 &880& 1000        \\
      \midrule
      DCT-RNN-GCN\cite{mao2020history}& 74.5 &89.0 &100.3 &108.2&108.2 &120.6 &135.9& 146.9&52.7 &57.8 &62.0 &64.9& 77.3 &91.8 &104.1 &112.1\\
      STS-GCN\cite{sofianos2021space} & 47.3 &56.8 &66.1 &72.0&74.7 &85.7 &96.2 &102.6&38.9 &44.0 &48.2 &51.1&50.8 &60.1 &68.9 &75.6\\
      \midrule
      Ours (L2D3) &\textbf{43.5}&	\textbf{54.5}&	\textbf{61.9}&	\textbf{68.8}&	\textbf{70.3}&	\textbf{82.0}&	\textbf{91.0}&	\textbf{97.7}&
      	\textbf{35.2}&	\textbf{41.9}&	\textbf{47.0}&	\textbf{50.0}&	\textbf{47.9}&	\textbf{58.6}&	\textbf{66.5}&	\textbf{72.9}
    \end{tabular}}
  \caption{ MPJPE error in mm for short-term prediction of 3D joint positions on Human3.6M. Our model (L2D3) outperforms the
  state-of-the-art by a small margin. We use 10 frames (400 msec) as input and predict 14-25 frames(560-1000 msec).}
\label{tab:2}
\end{table*}

\noindent
\begin{table}
  \centering
  \begin{tabular}{@{}lc|lc@{}}
    \toprule
    msec & \multicolumn{3}{c}{400}\\
    \midrule
    \multicolumn{2}{c|}{D = 4} & \multicolumn{2}{c}{L = 6} \\
    \midrule
    L = 2    &35.1                  &D = 1  &36.3\\ %8.9 35.1 73.6
    L = 3    &35.5                  &D = 2  &35.2 \\
    L = 4    &35.0                  &D = 3  &34.6 \\
    L = 5    &35.5                  &\textbf{D =4} &\textbf{34.4}\\
    \textbf{L=6}    &\textbf{34.4}  & D = 5  &37.1 \\
    L = 7     &34.8                 &D = 6  &35.8 \\
    L = 10     &35.2                &D = 7  &36.2 \\
    % L = 25    &     \\ 
  %   \midrule
  %   \multicolumn{2}{l}{L = 6} \\
  %   \midrule
  %   D = 1  &36.3 \\  
  %   D = 2  &35.2 \\  
  %   D = 3  &34.6 \\
  %   $\mathbf{D =4}$ &$\mathbf{34.4}$ \\
  %   D = 5  &37.1 \\ 
  %   D = 6  &35.8 \\
  %   D = 7  &36.2 \\
    \midrule
      \textbf{L=6, D=4}&\multicolumn{3}{c}{\textbf{34.4}}  \\
    \bottomrule
  \end{tabular}
  \caption{The influence of temporal receptive field L and graph max hop distance D to MPJPE Error. In the first part, 
  we fix the D = 4 and see the influence of L. In the second part, we fix the L = 6 and see the influence of D on the model.}
  \label{tab:LD}
\end{table}
\begin{table}
  \centering
  \begin{tabular}{@{}lccc@{}}
    \toprule
    msec & 80& 400 &1000\\
    \midrule
    no attention L2D3 & 9.1   &35.8&74.4\\
    attention L2D3  & 9.0 & 35.6&73.7  \\
    learnable graph L6D4  &9.0&49.8&96.5\\
    SAA-ar-MGCN L2D3&\textbf{7.6}&35.3&77.6\\
    SAA-2-MGCN L2D3       &8.5&36.0&77.8\\
    SAA-10-no refine L2D3     &9.1&36.1&76.4\\
    SAA-10-no refine L6D4     &9.0&35.3&82.2\\
    % \midrule
    SAA-10-MGCN L2D3      & 8.9            & 35.1         & \textbf{72.9} \\
    SAA-10-MGCN L6D4      & 8.6 & \textbf{34.4} & 83.6 \\
  %   sequence-aware attention_L25_D4      & $\mathbf{ }$  &$\mathbf{}$&$\mathbf{89.3}$ \\
  %   sequence-aware attention_L3_D4      & $\mathbf{ }$  &$\mathbf{}$&$\mathbf{75.8}$ \\
  %   sequence-aware attention_L3_D4_[16,16,16]      & $\mathbf{ }$  &$\mathbf{}$&$\mathbf{73.88}$ \\
  %   sequence-aware attention_L2_D4_[16,16,16]      & $\mathbf{ }$  &$\mathbf{}$&$\mathbf{72.8}$ \\
  %   sequence-aware attention_L2_D2_[16,16,16]      & $\mathbf{ }$  &$\mathbf{}$&$\mathbf{76.4}$ \\
    % sequence-aware attention_L1_D4_[16,16,16]      & $\mathbf{ }$  &$\mathbf{}$&$\mathbf{75.9}$ \\
    % sequence-aware attention_L1_D1_[16,16,16]      & $\mathbf{ }$  &$\mathbf{}$&$\mathbf{73.4}$ \\
    %sequence-aware attention_L10_D10      & $\mathbf{ }$  &$\mathbf{}$&$\mathbf{89.3}$ \\
    \bottomrule
  \end{tabular}
  \caption{MPJPE error in mm for prediction of 3D joint positions on Human3.6M. All are given 10 frames to 
  predict the 2, 10, 25 frames(80, 400, 1000 msec) in the future respectively. We experiment with the effects of the
  seq-attention mechanism, refine-MGCN, time receptive L and max hop D for the graph.
  No attention L2D3
  means not using attention and setting L =2, D = 3, and attention L2D3 means using normal attention and setting L
  =2, D = 3.
      }
  \label{tab:ablation}
\end{table}

\begin{table}[!htbp]
  \centering
  \begin{tabular}{@{}lccc@{}}
    \toprule
    msec & 80& 400 &1000\\
    \midrule
    H3.6M(STS-GCN\cite{sofianos2021space})& 10.1   &38.3&75.6\\
    H3.6M(our)& \textbf{8.6}   &\textbf{34.4}&\textbf{72.9}\\
    \midrule
    3DPW(STS-GCN\cite{sofianos2021space}) & 8.6   &24.5&42.3 \\
    3DPW(our) & \textbf{8.1}   &\textbf{24.0}&\textbf{40.3} \\
    \midrule
    AMASS(STS-GCN\cite{sofianos2021space}) & 10.0   &24.5&45.5 \\
    AMASS(our) & \textbf{8.0}   &\textbf{24.1}&\textbf{42.6} \\

  %   \midrule
  %   SAA-10-MGCN L2D3      & 8.9            & 35.1         & \textbf{72.9} \\
  %   SAA-10-MGCN L6D4      & \textbf{8.7} & \textbf{34.4} & 83.6 \\
    \bottomrule
  \end{tabular}
  \caption{Average MPJPE in mm. For the 80ms and 400ms prediction tasks, we set L=4 and D=6, 
  while for the 1000ms prediction task, we set L=2 and D=3. In particular, for the experiments 
  on AMASS and 3DPW datasets, we trained the model on AMASS and tested it on the BMLrub test 
  sequences of both AMASS and 3DPW.}
  \label{tab:experiment2}
\end{table}
\begin{figure}[!hbtp]
  \centering
     \includegraphics[width=0.8\linewidth]{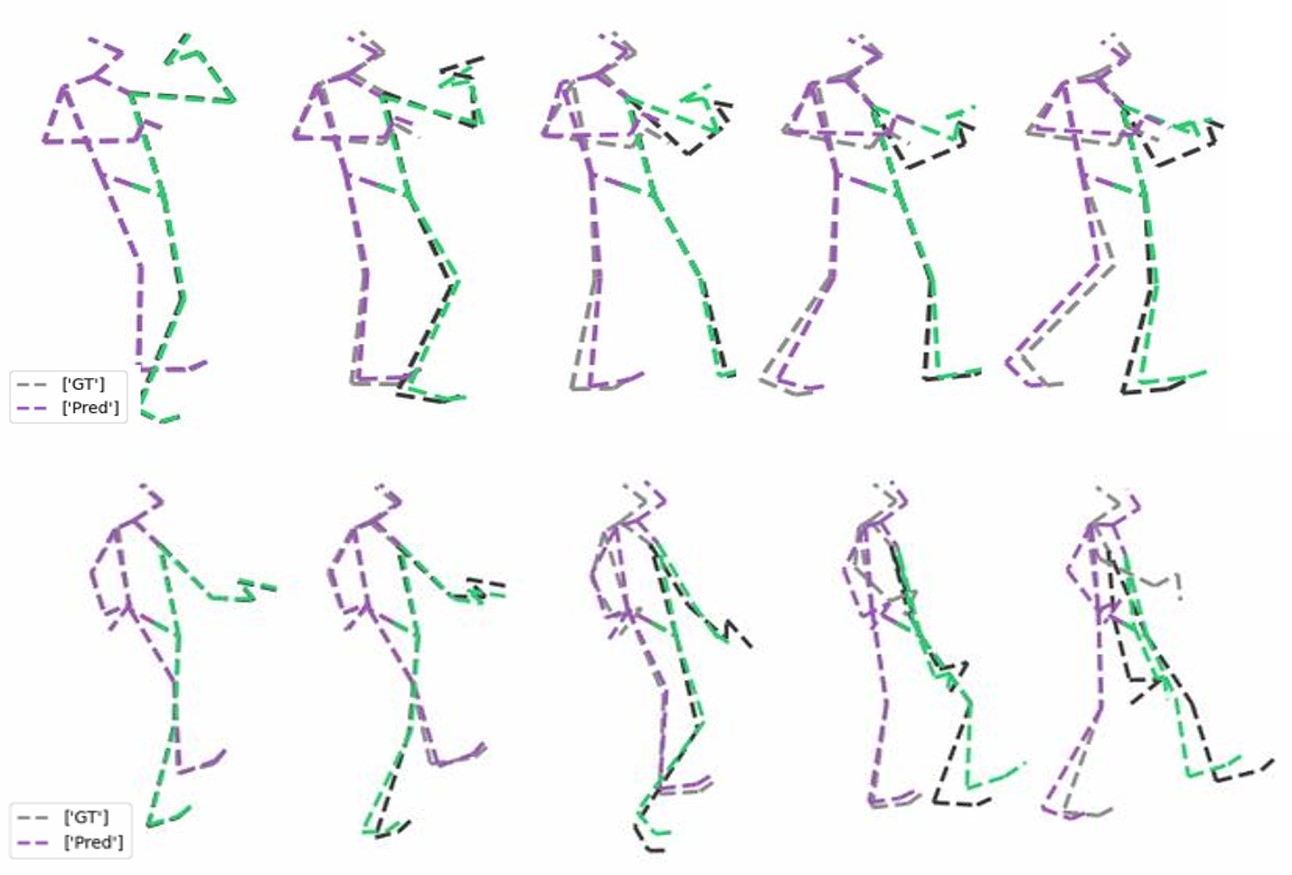}  
     \caption{Sample long-term predictions (25 frames, 1 sec) for the actions of Eating and Walking.
     Purple/green limbs are the left/right sides of the body. Gray/black pictorials
    indicate the observed ground-truth (GT) skeletons.}
  \label{fig:visualization}
\end{figure}

\subsection{Ablation study}
\label{sec:ab}

\textbf{Selection of L and D} 
We conducted experiments to investigate how the temporal receptive field length (L) and 
the maximum hop distance in a graph (D) impact the performance of our model.

% In \Cref{tab:ablation}, we examine the impact of L, D, 
% and sequence-aware attention mechanisms on the model's performance.
The \Cref{tab:LD} show that both L and D can negatively affect model performance if 
they are too short or too long.
By comparing the "*-L2D3"\footnote[1]{* refer to SAA-10-MGCN or SAA-10-no refine}
and the "*-L6D4" models in \Cref{tab:ablation}, we found that the model with L6D4 outperformed 
the model with L2D3 for short-term prediction tasks.

\textbf{Importance of natural link} 
To explore the impact of the natural link on our model, 
we conduct experiments on the model with a learnable graph where all joints interconnected.
% we only made the graph trainable. 

Comparing the "learnable graph L6D4" 
and "SAA-10-MGCN L6D4" models in \Cref{tab:ablation}, we observed a significant 
performance gap between the learnable graph and the natural link. This finding 
indicates that the natural connection plays a crucial role in human pose recognition.

\textbf{Sequence-aware attention} 
Comparing the "no attention L2D3" and "attention L2D3" models 
in \Cref{tab:ablation}, we found that 
models with attention outperformed those without attention on all prediction 
tasks. Furthermore,
comparing the "attention L2D3" and "SAA-10-MGCN L2D3" models,
sequence-aware attention performed better than normal attention.
% Although the model with normal attention (attention L2D3) performed better than 
% the SAA-10-MGCN L2D3 model on the 2 frames prediction task, the opposite was true 
% for long-term prediction tasks. 
Comparing "SAA-2-MGCN L2D3" and "SAA-10-MGCN L2D3", more anchors benefit the score in long-term task.
Notably, the "SAA-ar-MGCN" achieved the best performance 
on the 2 frames prediction task.
It illustrates Pseudo-autoregressive strategy performance better than anchor strategy in 2 frames task.
Therefore, we can choose different strategies for different problems to achieve the optimum.

% The experiments show that the two sequence-aware attention strategies mentioned 
% in Section \Cref{sec:3.5} improved the results to different degrees on different tasks.

\textbf{Refine MGCN}
Comparing the last four models in \Cref{tab:ablation}, we observed that the "refine model" 
significantly improved the model's performance for both short-term and long-term prediction tasks.

\subsection{Other Findings and Discussion}
\label{Sec:discuss}

% The experiments on the temporal dimension L and graph hop distance D highlighted 
% the importance of the natural structure of human joints. As shown in \Cref{tab:LD}, 
% the middle distance on the graph performed better on short-term prediction tasks. 
% This benefit stems from the deep neural networks and detailed graph descriptions. 
% However, on long-term prediction tasks, long-distance connections can cause serious 
% overfitting, as demonstrated in \Cref{tab:experiment}. 
% We experimentally show that proposed model

Further analysis reveals predicting changeable movements, such as discussions, is more challenging 
than normal movements that typically involve periodic motion, 
like walking (See \Cref{tab:1} and \Cref{tab:2}). 
Visual comparisons between our model and the ground truth are presented 
in \Cref{fig:visualization}.

The results in \Cref{tab:experiment2} demonstrate that our proposed approach 
is effective and generalizable for different datasets. Our model achieved superior 
performance compared to state-of-the-art methods in terms of short-term and long-term 
predictions of 3D coordinates for both AMASS and 3DPW datasets. These results further 
validate the robustness and effectiveness of our proposed approach in predicting human poses over time.

% The reason why GCN 
% \textbf{Discussion of MGCN}
GCN was adopted as a shape constraint model to avoid 
the physically impossible pose.
Our MGCN extends this effect in the time dimension: GCN can constrain the structure, 
and MGCN can extra constrain the joint trajectory. GCN is a sort of fully connected network with prior information 
describing the connections, and if the graph is learnable, the topology information will be lost.
We consider that the changeable motions are the most difficult task and 
may need to be offered more prior information like action category.
And predicting movement from key points ignores important appearance features.

\section{Conclusions}
This paper has proposed a novel approach called multi-graph convolutional network (MGCN). 
The key to the effectiveness of the framework is that we simultaneously extract the spatial-temporal 
information and decompose the task into many steps.
The proposed MGCN includes two attention strategies that make the model aware of the sequence, 
leading to state-of-the-art results on different tasks:
1) Pseudo-autoregressive strategy considers outputs as a recursive addition of the previous offsets;
2) Anchor strategy considers outputs as a convex combination of a set of anchors.
Furthermore, by incorporating a post-processing step called Refine-MGCN, we were able to achieve significant 
improvements in the results. These findings provide further evidence for the superiority of our MGCN model.
To build upon this approach, future work could consider applying it to other related 
problems such as human action recognition and pose interpolation. 

% \section{Acknowledgement}
% This work is supported by the 
% National Key Research and Development Program of China (2022YFC3602601).

%%%%%%%%% REFERENCES
{\small
\bibliographystyle{ieee_fullname}
\bibliography{MGCN}
}

\end{document}